\documentclass{article}

 \usepackage[final,nonatbib]{neurips_2023}

\usepackage[utf8]{inputenc} % allow utf-8 input
\usepackage[T1]{fontenc}    % use 8-bit T1 fonts
\usepackage{hyperref}       % hyperlinks
\usepackage{url}            % simple URL typesetting
\usepackage{booktabs}       % professional-quality tables
\usepackage{amsfonts}       % blackboard math symbols
\usepackage{nicefrac}       % compact symbols for 1/2, etc.
\usepackage{microtype}      % microtypography
\usepackage{xcolor}         % colors

\usepackage{comment}
\usepackage{amsmath}
\usepackage{graphicx}
\usepackage{hyperref}
\usepackage{float}
\usepackage{multirow}
\usepackage{makecell}
\usepackage{amsmath}
\usepackage{subcaption}
\usepackage{comment}
\usepackage{array}

\usepackage{caption}
\usepackage[title]{appendix}

\title{Generating Medication Prescriptions with Conditional Transformer} 
\author{%
Samuel Belkadi $^{\dagger}$, Nicolo Micheletti $^{\dagger}$, \textbf{Lifeng Han}, Warren Del-Pinto, \textbf{Goran Nenadic} \\ \vspace*{0.075cm}
             The University of Manchester\\ 
            {\tt \{samuel.belkadi, nicolo.micheletti\}@student.manchester.ac.uk} \\
            {\tt \{lifeng.han,warren.del-pinto,g.nenadic\}@manchester.ac.uk} \\
            $^\dagger$ \textit{Equal contribution} \\
}

\begin{document}

\maketitle
\begin{abstract}
Access to real-world medication prescriptions is essential for medical research and healthcare quality improvement. 
However, access to real medication prescriptions is often limited due to the sensitive nature of the information expressed. 
Additionally, manually labelling these instructions for training and fine-tuning Natural Language Processing (NLP) models can be tedious and expensive. 
We introduce a novel task-specific model architecture, Label-To-Text-Transformer (\textbf{LT3}), tailored to generate synthetic medication prescriptions based on provided labels, such as a vocabulary list of medications and their attributes.
LT3 is trained on a set of around 2K lines of medication prescriptions extracted from the MIMIC-III database, allowing the model to produce valuable  
%LT3 is trained on a vast corpus of medication prescriptions extracted from the MIMIC-III database, allowing the model to produce valuable  
synthetic medication prescriptions. 
We evaluate LT3's performance by contrasting it with a state-of-the-art Pre-trained Language Model (PLM), T5, analysing the quality and diversity of generated texts. 
We deploy the generated synthetic data to train the SpacyNER model for the Named Entity Recognition (NER) task over the n2c2-2018 dataset.
The experiments show that the model trained on synthetic data can achieve a 96-98\% F1 score at Label Recognition on Drug, Frequency, Route, Strength, and Form.
LT3 codes and data will be shared at \url{https://github.com/HECTA-UoM/Label-To-Text-Transformer}
\end{abstract}

%The proposed model seeks to mitigate the difficulties in healthcare data accessibility, bolstering medical research endeavours while ensuring the confidentiality of patient data.
%This study employs NLP techniques to overcome these challenges by generating synthetic medication prescriptions that can stand in for real-world data. 
% utilising the Transformer-based architecture, 

\section{Introduction}

Access to real-world medication prescriptions is pivotal for advancing medical research, including clinical natural language processing (NLP) applications, which is useful for improving healthcare quality and fostering the creation of novel solutions to address current research challenges \cite{NN22,alrdahi2023medmine,cui-etal-2023-medtem2}. 
However, given the confidential nature of these instructions, there are significant difficulties in acquiring and utilising them for research purposes  \cite{spasic2014textMine_cancer}. Additionally, manual labelling of such data for training and fine-tuning NLP techniques is labour-intensive and costly. This %aspect 
is also discussed by recent overview work in %the field 
\cite{wornow2023shaky_LLM4EHR}.

In response to these challenges, this study harnesses NLP methodologies to generate synthetic medication prescriptions. These synthetic examples provide a feasible alternative when real medical data is not available, which is a common problem due to concerns about patient confidentiality. The use of this synthetic data alongside, or in place of, real medical data can therefore alleviate challenges associated with accessing and employing sufficient data for NLP research, which is essential for healthcare quality enhancement and the inception of innovative strategies toward better computational modelling of digital healthcare data
%to address issues in computational modelling 
\cite{chen2019}.

The generation of synthetic clinical data has gained attention in recent years due to the challenges associated with accessing real-world clinical data \cite{Gonçalves2020, marchesi2022mitigating}.
Several studies have explored synthetic data generation for clinical NLP tasks. For instance, Amin-Nejad et al. \cite{amin-nejad-etal-2020-exploring} proposed a methodology for generating synthetic clinical text using structured patient information in a sequence-to-sequence manner and experimented with state-of-the-art Transformer models. They demonstrated that their augmented dataset could outperform baseline models on a downstream classification task.

Lee et al. \cite{DBLP:journals/corr/abs-1806-01353} explored the use of an encoder-decoder model to generate synthetic chief complaints from discrete variables in EHRs, such as age group, gender, and discharge diagnosis. After being trained end-to-end on authentic records, the model generated realistic chief complaint text that preserved the epidemiological information encoded in the original record-sentence pairs. 
This suggests that such a model could support the de-identification of text in EHRs, helping address the significant privacy concerns that often limit the sharing and use of real-world clinical data. 
However, only some works have attempted to control the generation of these models \cite{DBLP:journals/corr/abs-1909-05858}. Despite these advances, there is still room for improvement in generating synthetic clinical letters.

This study puts forth a novel task-specific model architecture, the Label-To-Text-Transformer (LT3), crafted to generate synthetic medication prescriptions. 
%Based on the Transformer's architecture \cite{vaswani2023attention} and trained on an expansive corpus of medication prescriptions, 
Based on the Transformer's architecture \cite{vaswani2023attention} and trained on an extracted set of around 2K medication prescriptions, 
LT3 is adept at generating high-quality synthetic medication prescriptions by capturing the unique patterns and dependencies involved in \textit{prescription writing} and other aspects of clinical documentation, such as sentence formatting. For example, given a medication "\textit{docusate sodium}" we would expect to generate a prescription such as "\textit{docusate sodium 100 mg Capsule Sig: One (1) Capsule PO BID (2 times a day) as needed for constipation.}".\\
To test how effective LT3 is, we will compare its performance to that of another State-of-the-art Pre-trained Language Model (PLM), T5 \cite{10.5555/3455716.3455856T5}, which we fine-tuned for this particular task.
For downstream applications, we also deploy the synthetic data generated by LT3 for training the SpacyNER model to compare the model performance with the ones trained from real data.

\section{LT3: Label-To-Text-Transformer}

\subsection{Problem Formulation}
\label{appendix:problem_formulation}
Let $\mathcal{C}$ be a space of clinical instruction features, and $c \in \mathcal{C}$ represents a feature vector for individual clinical instruction, e.g. a sentence piece. 
Let $\mathcal{L}$ be a set of drug labels. 
We have a dataset  $\mathcal{D}_C^L$ with labels annotated over the clinical instructions.
 
For each drug label $l \in \mathcal{L}$, we originally have a sub-set data $\mathcal{D}^{l}$ %$D_{l}$ 
defined as $\mathcal{D}^{l} = \{c_{n}^{l}\}^{N_{l}}_{n=1}$ %$D_{l} = \{c_{n}^{l}\}^{N_{l}}_{n=1}$ 
containing clinical instructions associated with drug $l$. Individual instructions are indexed by $n$ for each $l$, where $N_{l}$ is the number of instructions for drug $l$.

Our primary objective is to generate a synthetic dataset that replaces the real datasets entirely, conditioned on the drug labels from $\mathcal{L}$.
To achieve this, we aim to learn a density function $\hat{d}\{C|l\}$, which approximates the true distribution $d\{C|l\}$ of the clinical instructions conditioned on each drug label $l$.

Once the distributions for each drug label $l$ are learned, we generate an entirely synthetic dataset by drawing random variables from $\hat{d}\{C|l\}$ for each drug $l$. This synthetic dataset will have clinical instructions corresponding to every drug label in $\mathcal{L}$ and completely replace the original dataset.

\subsection{Model Architecture}

We introduce a transformer-based architecture, LT3 
%(problem formulated in Appendix \ref{appendix:problem_formulation}), 
with both an encoder and a decoder. The encoder processes the input labels, specifies drug names, and produces a contextualised representation, which is subsequently used by the decoder to generate output sequences in the form of prescriptions.

\begin{figure}[h]
  \centering
  \begin{minipage}[b]{0.76\textwidth}
    \includegraphics[width=\textwidth]{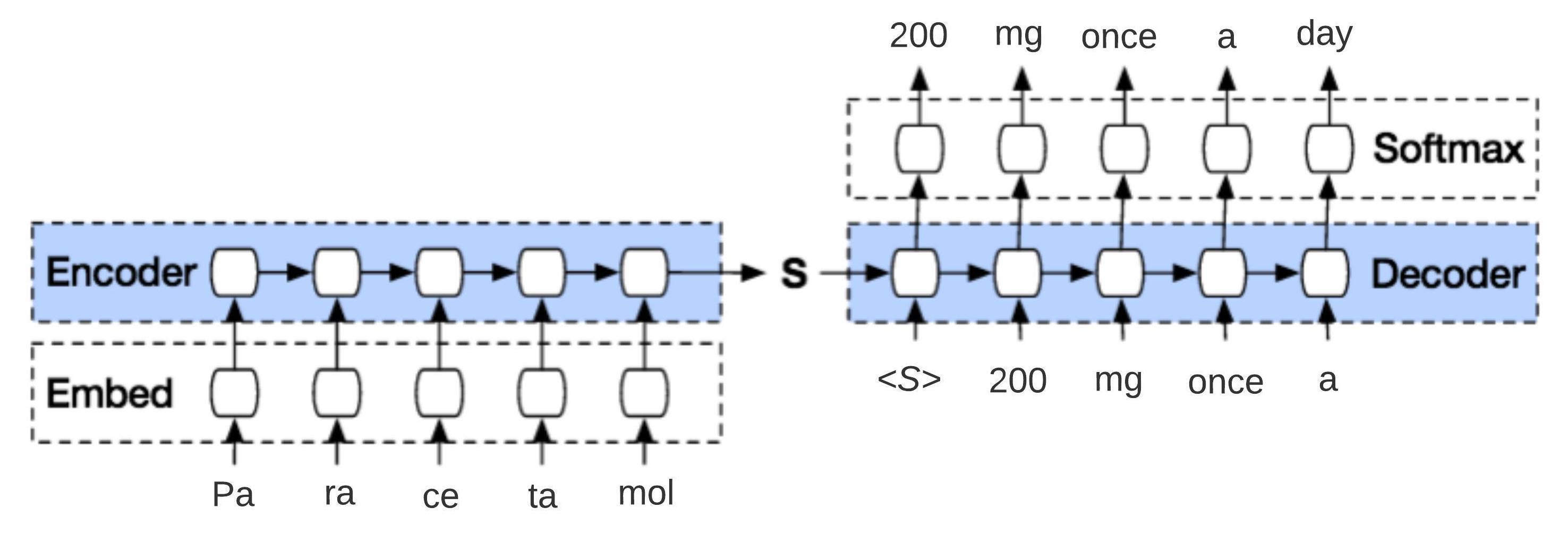}
    \caption{LT3 Architecture with input/output behaviour (this is a shortened example of a generated synthetic medical prescription.) %The generated dosage can change with the model settings. Please do not use this as medical instructions.
    }
    \label{fig:LT3_architecture}
  \end{minipage}
\end{figure}

LT3 implements the pre-trained word-piece BERT tokeniser \cite{devlin2019bert}. This selection is motivated by the objective of representing words as a series of smaller sub-word tokens. Simultaneously, this approach serves the dual purpose of minimising vocabulary size while handling unseen words as
the composition of  
a set of known sub-words.
% \subsubsection{Embedding layers}
Embedding layers are used within the model's architecture and are trained from \textit{scratch} to precisely cater to the requirements of the medical prescription writing task (Figure \ref{fig:LT3_architecture}).

%\begin{figure}[H]

\subsection{B2SD: Beam Search Decoding using Backtracking}
LT3 implements a novel Beam Search Decoding method using Backtracking (\textbf{B2SD}). While the conventional technique adopts a greedy strategy, selecting the best \textit{n} next-token candidates at each decoding step based on an overall probability function, this method instead employs a backtracking strategy \cite{10.1145/321296.321300_backtrack1965}.

At each step, we select the best candidate sequence generated so far. This selection relies on a heuristic function, specifically a joint probability function. Subsequently, the selected sequence is expanded by its best \textit{n} next-token candidates, referred to as a beam. This strategy allows the search tree to be flexible in size rather than limited to a fixed $n*seq_{len}$. However, in addressing the notable space and time complexity challenges of the B2SD algorithm, we decided to restrict the explorable space to the top-\textit{m} sequences generated so far, based on the same heuristic function. \\

In the example from Figure~\ref{fig:bsd_compare}, we compare the execution of both algorithms in generating sentences that describe someone as twelve years old. Both algorithms use a beam size of two and generate two sequences. The desired outputs are the ones with the highest total joint probabilities, namely "I am twelve" (p=0.138) and "You are twelve" (p=0.135). 
When comparing their execution, we observe that the backtracking algorithm \textit{(b)} explores seven vertices, including one dead-end labelled "scored" (coloured in blue), in contrast to the original algorithm \textit{(a)}, which only examines six vertices. However, in this scenario, the probabilities are sufficiently close to prevent a greedy algorithm, such as the original one, from catching the best overall sequences. Therefore, one of the two optimal solutions remains undiscovered, and instead, the dead-end labelled "scored" is greedily considered optimal by the original algorithm. However, B2SD managed to discover both desired outputs at the price of an additional vertex exploration.

\begin{figure}[H]
  \centering
  \begin{subfigure}{0.49\textwidth}
    \includegraphics[width=\linewidth]{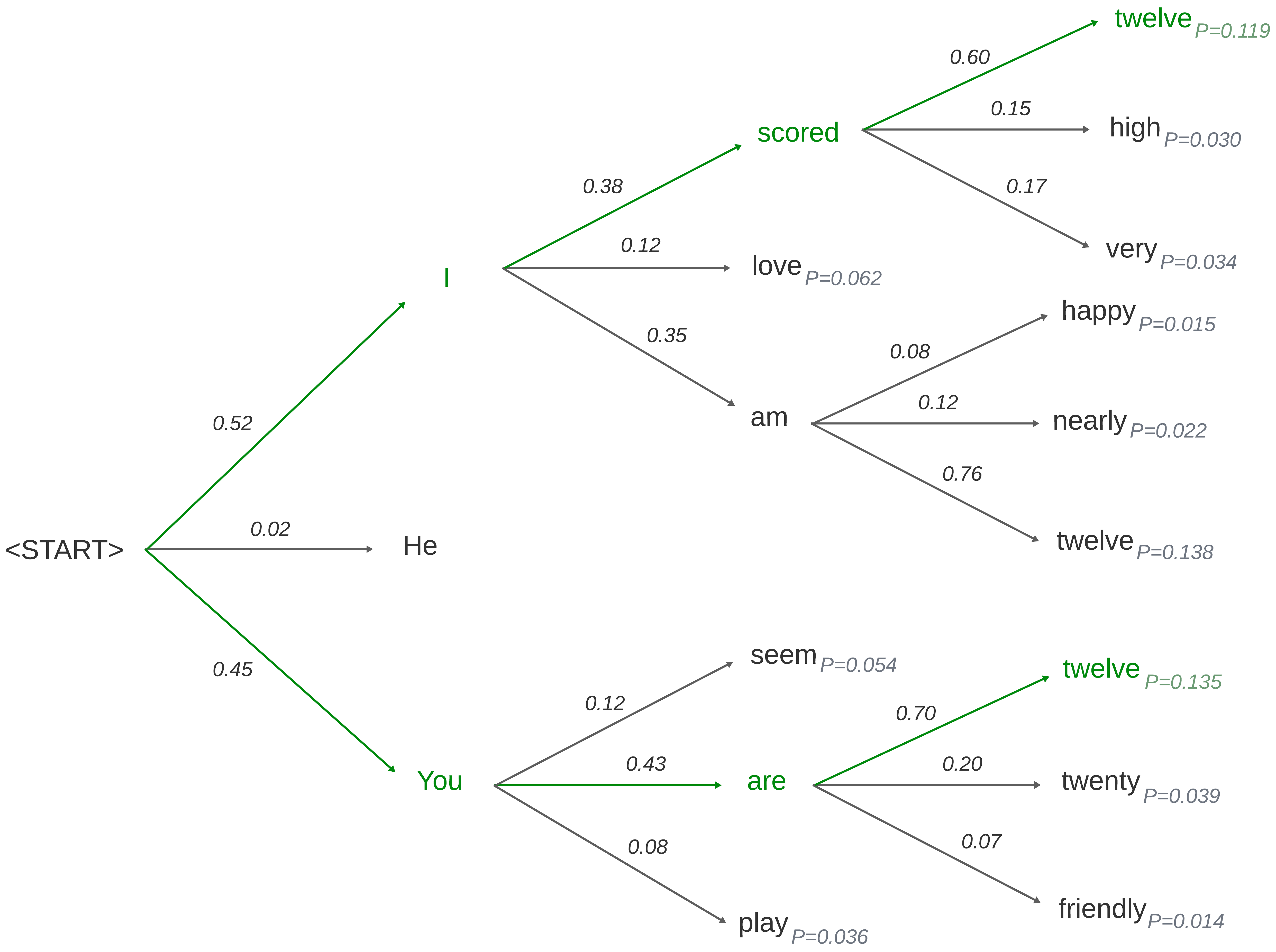}
    \caption{Original BSD ($n=2$)}
  \end{subfigure}
  \hfill
  \begin{subfigure}{0.49\textwidth}
    \includegraphics[width=\linewidth]{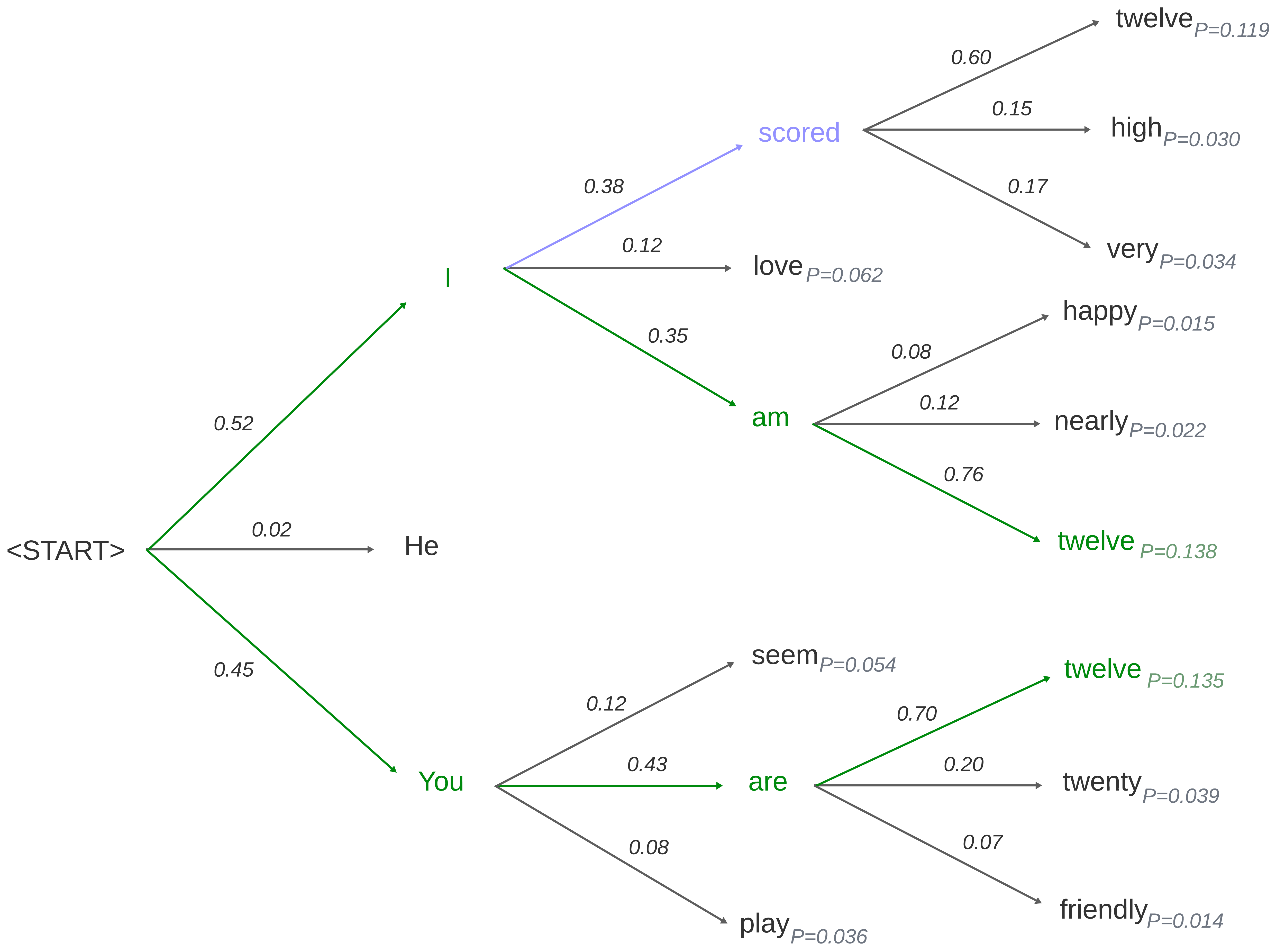}
    \caption{B2SD ($n=2$, $m=2$, $p_{b}=1$)}
  \end{subfigure}
  \caption{Execution Examples of Conventional Greedy BSD and B2SD Algorithms}
  \label{fig:bsd_compare}
\end{figure}

There is a trade-off between complexity and the main advantage of the backtracking algorithm, which is its ability to find the best solution in the beam tree according to its heuristic within a finite time compared to the original BSD algorithm. This means that a higher level of complexity may lead to a longer search time but a better solution. In our specific scenario, striking this balance is justified. That is because LT3 deals with a limited number of samples to generate relatively short sequences.
Moreover, by utilising this algorithm, we can efficiently bypass tokens within the beam that, while still within the top-n candidates, are significantly less likely to contribute to genuinely interesting sequences. This approach encourages the model to prioritise the development of promising sequences.

Therefore, the complexity of the newly proposed B2SD algorithm can be expressed as exponential in the sequence's length, denoted $\mathcal{O}(n^{seq_{len}})$. At the same time, the original one is linear: $\mathcal{O}(n*{seq_{len}})$. However, worst-case complexity may not represent the execution times for the above reasons (see Appendix \ref{appendix:beamsearchcomparisons}). %Figure ~\ref{fig:bsd_exec_times}).

Besides using this backtracking approach, the beam size \textit{n} does not need to be greater or equal to the number of desired output sequences. Instead, \textit{m} should follow this requirement, as it is the maximum number of sequences considered for output.

To enhance the quality of sequence generations, we implement an additional unigram repeat penalty targeting subsequences of length 4. This penalty aims to discourage the generation of sequences where a subsequence of four tokens contains multiple instances of the same token. For example, the subsequence [43, 32, 21, 43] incurs a penalty as the token "43" appears twice. The penalty itself is calculated using the following formula.

\begin{equation}
    p'(Y) = p(Y)^{2 - 0.5*p_{T}}
\end{equation}

where $p_{T}$ is the probability (or certainty) of the last duplicate token, here "43", and $p(Y)$ is the joint probability of the sequence $Y$. This design allows the application of a penalty that accounts for the token's certainty level. In cases where a duplicate token is suggested but has a high certainty, the penalty is reduced, considering that the model may intentionally repeat it to convey specific information. This can be the case in sentences such as "(once a day (at bedtime\textbf{))}" where closing parenthesis are repeated consecutively.

Finally, to further reduce the search space, the maximal probability difference in beam, $p_{b}$, constrains the tokens considered in a beam. This value tells how much lower the probability of a token in the beam from the top probability token in that same beam is allowed to be. For example, if the top token of a beam has a probability of 0.5 and $p_b = 0.5$, tokens in the beam with a probability $< 0.5*0.5$ won't be further considered. This is useful whenever an obvious best candidate exists, for instance, when selecting the drug name that was itself given as input.

Therefore, the beam size \textit{n}, maximum candidates space \textit{m}, and maximal probability difference in beam $p_{b}$ are three hyper-parameters to fine-tune to obtain optimal results. We assign them the values $n = 4\text{, } m = 3*nb_{output} \text{ and } p_{b} = 1$.
\\\\
\textbf{Heuristic function}\\
The heuristic function used is logarithmic in the sequence's joint probability 
\begin{equation}
    h(Y) = \frac{log_e(p(Y_{0,...,n}))}{lp(Y)}
\end{equation}

where $Y_{n}$ is the $n^{th}$ token of the sequence $Y$ generated so far, and $Y_{0,...,n}$ refers to the product of the probabilities associated with each token in the sequence $Y$, which is referred to as the joint probability of $Y$. 
The heuristic function applies length normalisation as taken from Google's NMT System paper \cite{johnson-etal-2017-googlesNMT}, where we set $\alpha = 0.6$. % this one? https://arxiv.org/abs/1611.04558 Google's multilingual NMT

\begin{equation}
    lp(Y) = \frac{(5 + |Y|)^\alpha}{(5 + 1)^\alpha}
\end{equation}

\section{Evaluation}

\subsection{Dataset and Preprocessing}

Our research draws upon a specialised subset of the MIMIC-III (Medical Information Mart for Intensive Care) database \cite{Johnson2016-tr, Johnson2020-ik}; specifically, the portion that aligns with the National NLP Clinical Challenges (n2c2) 2018 shared task data on adverse drug events and medication extraction with gold labels \cite{Henry2019} (Appendix \ref{appendix:n2c2data}).
%on adverse drug events and medication extraction in electronic %We chose the n2c2 dataset for two main reasons. First, it contains many caregiver notes and medication prescriptions over a varied range of clinical conditions and treatments, ensuring a broad spectrum of clinical letters can be generated by our models, enhancing their utility in different clinical scenarios. Second, the n2c2 dataset annotations conform to the 2010 i2b2/VA Challenge on Concepts, Assertions, and Relations in Clinical Text, a well-established and comprehensive framework for processing and understanding clinical text. This standardisation facilitates handling clinical notes' diverse and complex language patterns. Moreover, using these gold labels helps us ensure the accuracy and consistency of our model's learning process, which is crucial to generating high-quality synthetic clinical letters. In addition, using a dataset that adheres to a widely accepted annotation guideline enhances the replicability and validity of our study. It allows other researchers and practitioners to understand the method and results of our work within a known context, promoting transparency and further collaboration.
We divided the official training set into our "training" and "validation" sets with the ratio (9:1) and kept the original test set.
We implemented a procedure in our dataset to automatically extract and structure discharge medication information from the n2c2 dataset. The procedure scans each text-based medical record in the original dataset and identifies the text segment containing information about the medications prescribed upon discharge. 

The identified medication data is further decomposed into two primary components: the label (or name of the medication) and the associated instructions. Both are captured and stored in a structured format. Finally, we apply statistical filtering techniques to remove outliers based on the medication labels' length and instructions. This ensures a dataset free from extreme values that could potentially bias downstream applications.

\subsection{Model Selection}

We conduct a model evaluation experiment to select the most optimal LT3 model (Appendix \ref{appendix:model_selection}). This experiment entails training each model on the training set and using them to generate five times the amount of data from the validation set as synthetic data. We then assess the models' performance using the quantitative metrics BLEU, ROUGE-1/2/L, and BERTScore. Based on the results, we select the best model and retrain it on the training and validation sets to obtain a final LT3 model.

\begin{comment}
    \begin{figure}[H]
  \centering
  \begin{minipage}[b]{1\textwidth}
    \includegraphics[width=\textwidth]{images/model_selection.png}
    \caption{Model Selection Pipeline}
    \label{fig:model_selection_pipeline}
  \end{minipage}
\end{figure}
\end{comment}

For the T5 model, given the provided labels, we leverage T5 language processing capabilities to fine-tune the model to generate appropriate text responses in the form of medication prescriptions from labels representing medications such as "paracetamol" or "ibuprofen".

\subsection{Lexical Similarity Evaluation against References}

For this experiment, we fine-tuned three versions of T5, namely t5-small, t5-base, and t5-large, paired with their sentence-piece pre-trained tokeniser.
Each is fine-tuned independently on the same dataset as LT3 to provide comparable results, with the prompt "summarise:" as it is the closest to our task. The results in Table \ref{table:evaluation_metrics_model_evaluation} show that LT3's generations are the closest match to the reference samples. We use multi-reference evaluation to consolidate our results.
Refer to Appendix \ref{appendix:eval_settings} for more details on this evaluation's strategies and motivations.

\begin{table}[H]
  \centering
  \caption{Quantitative evaluation of LT3 (learned-scratch) vs T5 (fine-tuned) on the Testing Set. %\textit{\textbf{1st task:} for current split of train-valid: calculate the unseen medication numbers and ratios between (validation, train) and (test, train+validation)}. \textbf{2nd task:} can we run cross-validation, each time taking out a different portion of 10 per cent of the training set; this can be used to see if the evaluation on validation set score ranges, e.g. 60\% to 80\%? this will cover the testing score on test set 70 plus
  }
  \begin{tabular}{|c||c|c|c|c|c|}
      \hline
      Models & BLEU & ROUGE-1 & ROUGE-2 & ROUGE-L & BERTScore \\
      \Xhline{2\arrayrulewidth}
      T5 Small   & 71.75 & 76.16 & 66.24 & 75.55 & 0.70 \\
      T5 Base    & 71.98 & 76.28 & 66.30 & 75.45 & 0.70 \\
      T5 Large   & 69.89 & 75.07 & 65.19 & 74.22 & 0.68 \\ \hline
      LT3        & \textbf{78.52} & \textbf{78.16} & \textbf{68.72} & \textbf{77.55} & \textbf{0.72} \\
      \hline
  \end{tabular}
\label{table:evaluation_metrics_model_evaluation}
\end{table}

\subsection{Lexical Diversity Evaluation within Generated Outputs}

A diverse range of content is crucial in the note-generation process to create unbiased and individualised clinical instructions. To achieve this, we have implemented a diversity score that measures the breadth of our model's output. For each label, we measured the Jaccard similarity \cite{jaccard1908nouvelles,ivchenko1998jaccard}
%\footnote{\url{https://www.sciencedirect.com/topics/computer-science/jaccard-similarity}} 
score of the generations of our models.
A higher Jaccard Score indicates more similarity between the two populations. A lower score indicates better diversity in our tasks.
The results in Table \ref{table:jaccard_scores} show a lower intra-similarity score for the generations of LT3, implying that LT3 produces more diverse samples.

%\begin{table}[H]
\begin{table}[h]
    \centering
    \caption{Jaccard scores of LT3 and T5 on the testing set (lower score is better). }
    \begin{tabular}{|c||c|c|}
         \hline
          & Median Jaccard Score & Average Jaccard Score \\
         \Xhline{2\arrayrulewidth}
         LT3        & \textbf{ 0.650} & \textbf{0.652} \\
         T5 Base    & 0.658         & 0.660          \\
         \hline
    \end{tabular}
    \label{table:jaccard_scores}
\end{table}

\subsection{Downstream NER Task}

\begin{figure}[H]
  \centering
  \begin{minipage}[b]{0.9\textwidth}
    \includegraphics[width=\textwidth]{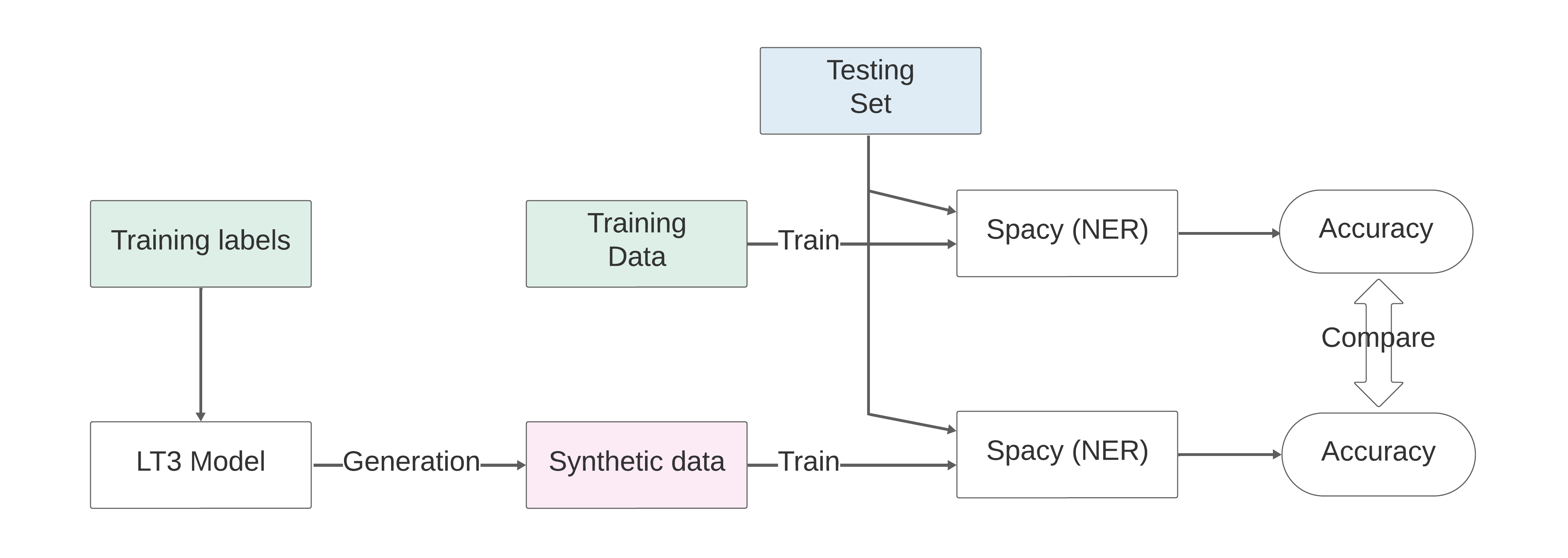}
    \caption{Cross-model Evaluation Pipeline}
    \label{fig:ner_pipeline}
  \end{minipage}
\end{figure}

In the cross-model evaluation (Figure \ref{fig:ner_pipeline}), we aim to substantially increase the size of our dataset beyond what we initially extracted from n2c2. To achieve this, we generate synthetic data using LT3 on the known training labels. This synthesis allows us to create a dataset that is five times larger than the original one.
Subsequently, we perform fine-tuning on Spacy\footnote{\url{https://spacy.io}} using both the original and synthetically generated datasets. Finally, we compare the three resulting NER models, one fine-tuned on the real dataset, one on the synthetic dataset, and the last on a combination of real and synthetic data. Specifically, the real dataset is oversampled, ranging from 100\% (identical to the original) to 500\% (five times the original size).
The synthetic dataset is generated using real labels, ranging from 100\% to 500\%.
The combined real and synthetic dataset starts with 100\% real data, to which synthetic data is incrementally added, from 100\% to 400\%.
The NER model is trained to recognise medical labels: Drug, Strength, Form, Route, and Frequency. 
This comparison helps us to quantify the effectiveness of using synthetic data generated using LT3 to augment or replace the training dataset by assessing the ability of the fine-tuned models to recognise named entities in unseen data. 

\begin{figure}[H]
    \centering
    \includegraphics[width=0.9\textwidth]{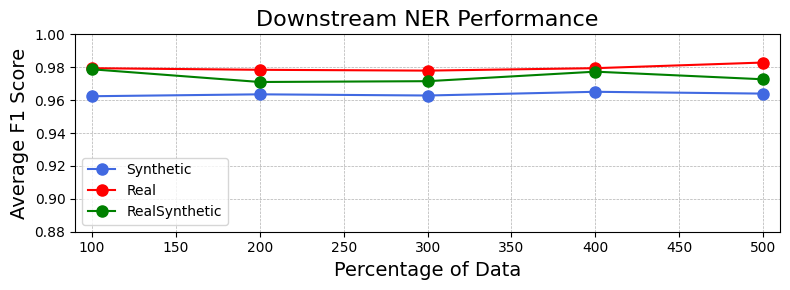}
    \caption{Average F1 score for five labels (Drug, Strength, Form, Route, Frequency) using Synthetic data, Real data, and Real+Synthetic. 
    RealSynthetic: 100\% real + n*100\% Synthetic. Real: over-sampled. %Drugs label always gets around 1.00 F1 score. %On Synthetic data: vertical scores on four labels - form, frequency, route, strength. Drugs always get an F score of 1. Horizontal line: size wise to 500\% times.
    %In comparisons, on real data, the best is 0.9686 with the same setting of 4 labels. \textbf{new task: }\textit{with adding more synthetic data to real data, how the eval score changes}.
    }
    \label{fig:NER-real-vs-syn-RealSyn}
\end{figure}

%\textit{In comparisons, on real data, the best is 0.9686 with the same setting of 4 labels.}
The evaluation scores F1 in Figure \ref{fig:NER-real-vs-syn-RealSyn} show that LT3 could successfully train Spacy on this NER task on five labels "drug, form, frequency, route, and strength" achieving 0.96+ scores. The evaluation on Drug labels always yields around 1.00 accuracy. Most importantly, it yielded comparable performance to the real data, demonstrating the quality of generated texts and the benefit of using the generated synthetic data as an alternative to real data.

\section{Conclusion and Future Work}
To facilitate clinical NLP research and address the data privacy and restriction issues, we proposed LT3 for generating synthetic clinical data using pre-defined drug labels and related attributes from the n2c2-2018 shared task. 
The evaluation against the T5 model demonstrated that LT3 can generate better quality and diversity outputs. 
Furthermore, utilising synthetic data generated by LT3 for the NER task demonstrated its ability to effectively train SpacyNER, resulting in performances comparable to those achieved with real data. This underscores the advantages of employing LT3 as a viable alternative to real data.
In future work, we plan to design new benchmarks on clinical NLP tasks using synthetic data to move the field forward. We also plan to conduct model training on new label sets such as "diagnoses" and generating full clinical letters.

 \section*{Author Contributions}
 SB and NM co-developed LT3, fine-tuned T5, and built an evaluation pipeline. Specifically, SB developed the LT3 architecture, code, and B2SD, and NM fine-tuned Spacy and deployed LT3's generated data on the NER.
 SB did closeness to reference evaluation; NM extracted+processed the dataset and implemented and ran intra-similarity evaluation.
 LH and GN designed the project and supervised the progress, and LH revised the first manuscript. 
 WDP co-supervised the project and revised the final manuscript.
 Everyone approved the final manuscript.

 \section*{Acknowledgements}
We thank Ms. Wuraola Oyewusi, Dr Christopher J. Hyde and anonymous reviewers for their valuable discussions and insightful comments on this project and the earlier manuscript. 
% .
SB and NM were partially supported by the University of Manchester student summer project via the Department of Computer Science.
LH, WDP, and GN are grateful for the support from the grant “Assembling the Data Jigsaw: Powering Robust Research on the
Causes, Determinants and Outcomes of MSK Disease”. The project has been funded by the Nuffield
Foundation, but the views expressed are those of the authors and not necessarily the Foundation. 
Visit www.nuffieldfoundation.org. LH, WDP, and GN were also supported by the grant “Integrating hospital outpatient letters into the healthcare data space” (EP/V047949/1; funder: UKRI/EPSRC).

%LH, WDP, and GN are grateful to Grant ``Assembling the Data Jigsaw: Powering Robust Research on the Causes, Determinants and Outcomes of MSK Disease'' (The project has been funded by the Nuffield Foundation, but the views expressed are those of the authors and not necessarily the Foundation. Visit www.nuffieldfoundation.org) and Grant EP/V047949/1 ``Integrating hospital outpatient letters into the healthcare data space'' (funder: UKRI/EPSRC).

%\section*{References}

\bibliographystyle{vancouver}
\bibliography{sample}

\clearpage

\begin{appendices}

\section{On Current PLMs for Clinical NLP}

Natural language processing (NLP) technologies have been increasingly used in healthcare over the past several years, contributing to advancements in several areas such as clinical decision support, patient triage, and automated clinical documentation \cite{Yang2022, Casey2022}. However, these applications face numerous challenges, one of the most significant being the scarcity of available data. This issue is predominantly due to stringent privacy regulations and the sensitive nature of healthcare data, which prevent access to large volumes of real-world clinical data \cite{Ive2020, Chapman2011OvercomingBT}.

To circumvent this problem, synthetic data generation has been explored as an alternative approach, aiming to produce data that mimics the properties and structure of real-world clinical data without compromising patient privacy \cite{10.1093/jamia/ocab112}. Despite this approach's potential, producing high-quality, domain-specific synthetic data remains challenging due to the complexity and specificity of medical language.

Pre-trained Language Models (PLMs) have shown remarkable capabilities in generating contextualised texts, such as translations \cite{DBLP:journals/corr/abs-2010-11934} and summaries \cite{SUM}. However, they have struggled to generate coherent text in the medical domain. This is due to the considerable shift from standard NLP tasks to the medical domain, which presents challenges as pre-trained models have a more general-purpose design and do not learn directly from restricted domain-specific data \cite{grambow-etal-2022-domain}. For example, the word "paracetamol" may be captured in many training documents that do not correspond to synthetic clinical letter generation tasks and, therefore, be a noisy contribution. Moreover, PLMs need more flexibility to handle different input types and are not explicitly trained on label-to-text data, resulting in sub-optimal accuracy for the specific task. To address these challenges, this research proposal aims to develop a task-specific model architecture that can overcome the limitations of pre-trained models and generate high-quality synthetic clinical instructions.

\begin{comment}
    \section{Problem Formulation}
\label{appendix:problem_formulation}
Let $C$ be a space of clinical instruction features, and $c \in C$ represents a feature vector for individual clinical instruction, e.g. a sentence piece. 
Let $L$ be a set of drug labels. 
We have a dataset  $\mathcal{D}_C^L$ with labels annotated over the clinical instructions.
 
For each drug label $l \in L$, we originally have a sub-set data $D^{l}$ %$D_{l}$ 
defined as $D^{l} = \{c_{n}^{l}\}^{N_{l}}_{n=1}$ %$D_{l} = \{c_{n}^{l}\}^{N_{l}}_{n=1}$ 
containing clinical instructions associated with drug $l$. Individual instructions are indexed by $n$ for each $l$, where $N_{l}$ is the number of instructions for drug $l$.

Our primary objective is to generate a synthetic dataset, conditioned on the drug labels from $L$, that replaces the real datasets entirely.

To achieve this, we aim to learn a density function $\hat{d}\{C|l\}$, which approximates the true distribution $d\{C|l\}$ of the clinical instructions conditioned on each drug label $l$.

Once the distributions for each drug label $l$ are learned, we generate an entirely synthetic dataset by drawing random variables from $\hat{d}\{C|l\}$ for each drug $l$. This synthetic dataset will have clinical instructions corresponding to every drug label in $L$ and completely replace the original dataset.
\end{comment}

\section{On the Choice of N2C2 Data}
\label{appendix:n2c2data}
We chose the n2c2 dataset for two main reasons. First, it contains many caregiver notes and medication prescriptions over a varied range of clinical conditions and treatments, ensuring a broad spectrum of clinical instructions can be generated by our models, enhancing their utility in different clinical scenarios. Second, the n2c2 dataset annotations conform to the 2010 i2b2/VA Challenge on Concepts, Assertions, and Relations in Clinical Text, a well-established and comprehensive framework for processing and understanding clinical text. This standardisation facilitates handling clinical notes' diverse and complex language patterns. Moreover, using these gold labels helps us ensure the accuracy and consistency of our model's learning process, which is crucial to generating high-quality synthetic medical data. In addition, using a dataset that adheres to a widely accepted annotation guideline enhances the replicability and validity of our study. It allows other researchers and practitioners to understand the method and results of our work within a known context, promoting transparency and further collaboration.

\section{On Evaluation Metrics}

BLEU \cite{papineni-etal-2002-bleu}, ROUGE \cite{lin-2004-rouge}, and BERT Score \cite{zhang2020bertscore} represent key evaluation metrics, each illuminating different facets of text quality. BLEU focuses on the syntactic elements, measuring the overlap of n-grams between the machine-generated text and a reference. It incorporates a brevity penalty for translation length, making it particularly useful for tasks like machine translation.

On the other hand, ROUGE (Recall-Oriented Understudy for Gisting Evaluation) is more recall-focused and assesses the quality of summaries by comparing them to reference summaries. It considers the number of overlapping units, such as n-grams, word sequences, and word pairs between the generated and reference summaries.

Finally, the BERT Score leverages the power of pre-trained language representations to go beyond mere syntactic overlap, capturing semantic nuances between predicted and reference texts through cosine similarity measures. These approaches reflect a shift from rigid, rule-based evaluations toward more dynamic, context-aware metrics, aligning more closely with human perceptions of text quality.\\ 

\section{Model Selection in Details}
\label{appendix:model_selection}

\subsection{On tokenisation}

Experiments were conducted to select the most effective tokenisation strategy for this task, for which results are summarised in Table~\ref{table:evaluation_metrics_model_selection} and Figure~\ref{fig:eval_training_loss}. 
Three different types of tokenisers were considered: a custom full-word tokeniser, a pre-trained word-piece tokeniser (BERT-base-cased), and a pre-trained sentence-piece tokeniser (T5-base).

Throughout the experiment, LT3 encountered challenges implementing the full-word tokeniser built from scratch. Although this tokeniser yielded overall good performances, it struggled with handling unknown words, for which the only solution seemed to be significantly expanding the vocabulary size to cover a vast tokenisation space. Without an extensive vocabulary, the tokeniser fails to map unseen words, leading to a lack of contextual understanding for LT3.

On the other hand, significant improvements were observed when using the word-piece tokeniser (BERT) due to his ability to represent any word as a sequence of smaller sub-words while minimising its vocabulary size. This allows the model to effectively handle unseen words and cover a large tokenisation space to yield better generalisation capabilities.

Experiments were also carried out using the pre-trained sentence-piece tokeniser provided by T5. This tokeniser demonstrated improvements similar to those of the word-piece tokeniser (BERT), effectively mitigating the issues faced by the custom tokeniser. However, we observed that the word-piece tokeniser (BERT) could generate predictions for unseen data at an earlier stage of training compared to the pre-trained sentence-piece tokeniser (T5). This might be due to LT3 generating short sentences with low correlation and no repetitive patterns between words, a task for which word-piece tokenisers may be more adapted.

Considering these factors, we concluded that the BERT word-piece tokeniser aligned most effectively with our task.

\subsection{On Embeddings}

Alternatively, this study explored two interesting embedding methods: transfer learning using pre-trained embeddings and embedding layers trained from scratch. 
Transfer learning used BioBERT (base-v1.1) embeddings, pre-trained on large medical corpora, including PubMed 1M, while embedding layers were trained during LT3's training phase.

Although transfer-learning can provide a solid foundation for the model, especially when task-specific data is scarce or when the pre-training domain closely matches the task, its experimental results displayed challenges when applied to our task (Table~\ref{table:evaluation_metrics_model_selection}). Despite training in medical texts, pre-trained embeddings could not grasp the prescriptions' nuances and unique formats. 
This led to a need for extensive training to overwrite the previous embeddings, as seen in Figure~\ref{fig:eval_training_loss}. 
On the other hand, embedding layers outperformed pre-trained embeddings by addressing the task's unique format and leveraging the extensive available data. As a result, LT3 displayed a much better learning shape and evaluation results when implementing embedding layers.\\

Note that, when using pre-trained embeddings, the disparity between the learning curve, which appears to be reasonably good (Table~\ref{table:evaluation_metrics_model_selection}), and the evaluation scores, which are rather very low (Figure~\ref{fig:eval_training_loss}), is attributed to the application of teacher forcing during training. This explains that the model with pre-trained embeddings can accurately predict the next token, provided with an accurate context and a generated sequence. However, it struggles when tasked with independently creating an appropriate context from the input and generating a complete sequence that is contextually coherent.

\subsection{Results}

\begin{figure}[]
  \centering
  \begin{minipage}[b]{0.8\textwidth}
    \includegraphics[width=\linewidth]{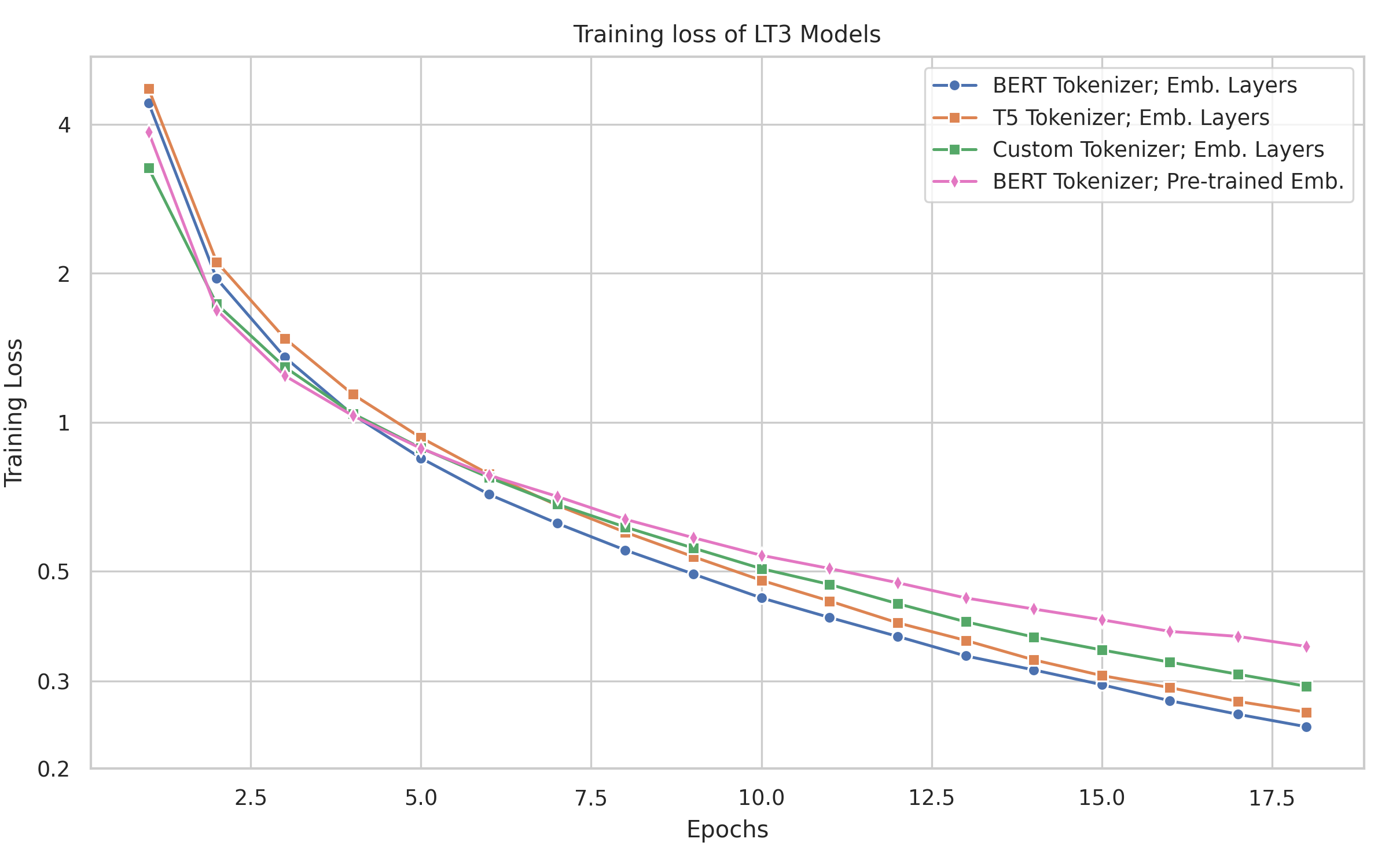}
    \caption{Training Loss of LT3 Models}
    \label{fig:eval_training_loss}
  \end{minipage}
\end{figure}

We plot the training loss (Figure \ref{fig:eval_training_loss}) and evaluation scores (Table \ref{table:evaluation_metrics_model_selection} on the validation set and Figure \ref{fig:bsd_compare} on the test set) to provide a comprehensive assessment of each model's learning trajectory and generation quality. This approach helps readers understand how each model evolves through the learning process.

%\subsubsection{Results}

\begin{table}[H]
  \centering
    \caption{Quantitative Evaluation Results of LT3 Models on the Validation Set}
    \begin{tabular}{|ccc||c|c|c|c|c|}
      \hline
      Tokenizer & Embeddings & Beam Search & BLEU & ROU-1 & ROU-2 & ROU-L & BERTScore \\
      \Xhline{2\arrayrulewidth}
      BERT     & Emb. layers & B2SD      & \textbf{66.31} & \textbf{70.74}  & \textbf{60.01} & \textbf{70.03} & \textbf{0.65} \\
               & Pre-trained &         & 36.11 & 43.16 & 28.56 & 41.81 & 0.29 \\
               & Emb. layers & Default & 54.33 & 67.01 & 55.46 & 66.20 & 0.60 \\
      
      Custom   &             & B2SD      & 64.19 & 70.00 & 58.34 & 68.13 & 0.63 \\
      T5-base  &             &         & 65.78 & 68.99 & 58.63 & 68.22 & 0.63 \\
      \hline
    \end{tabular}
    \label{table:evaluation_metrics_model_selection}
\end{table}

\begin{figure}[]
  \centering
  \begin{subfigure}{0.49\textwidth}
    \includegraphics[width=\linewidth]{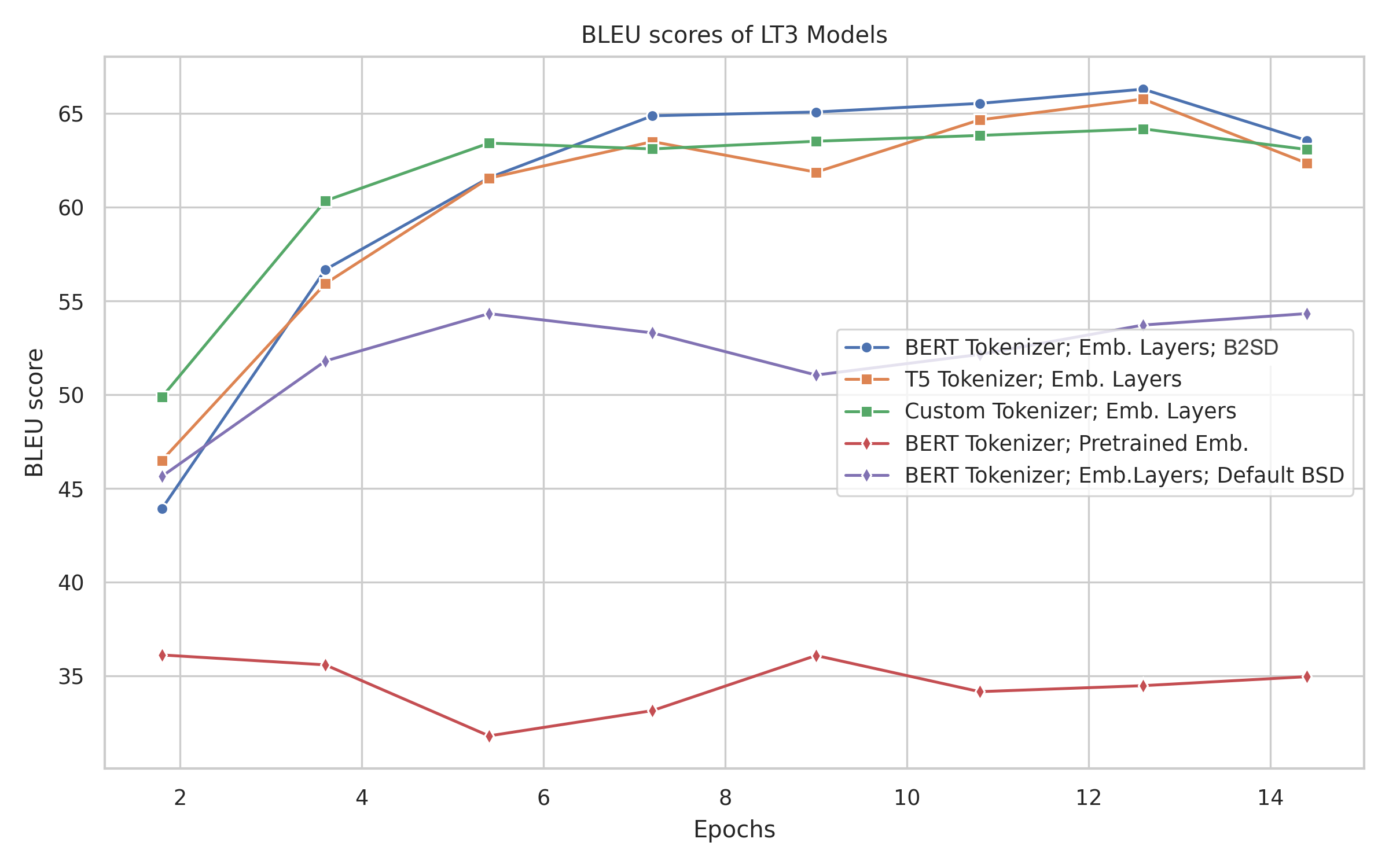}
  \end{subfigure}
  \hfill
  \begin{subfigure}{0.49\textwidth}
    \includegraphics[width=\linewidth]{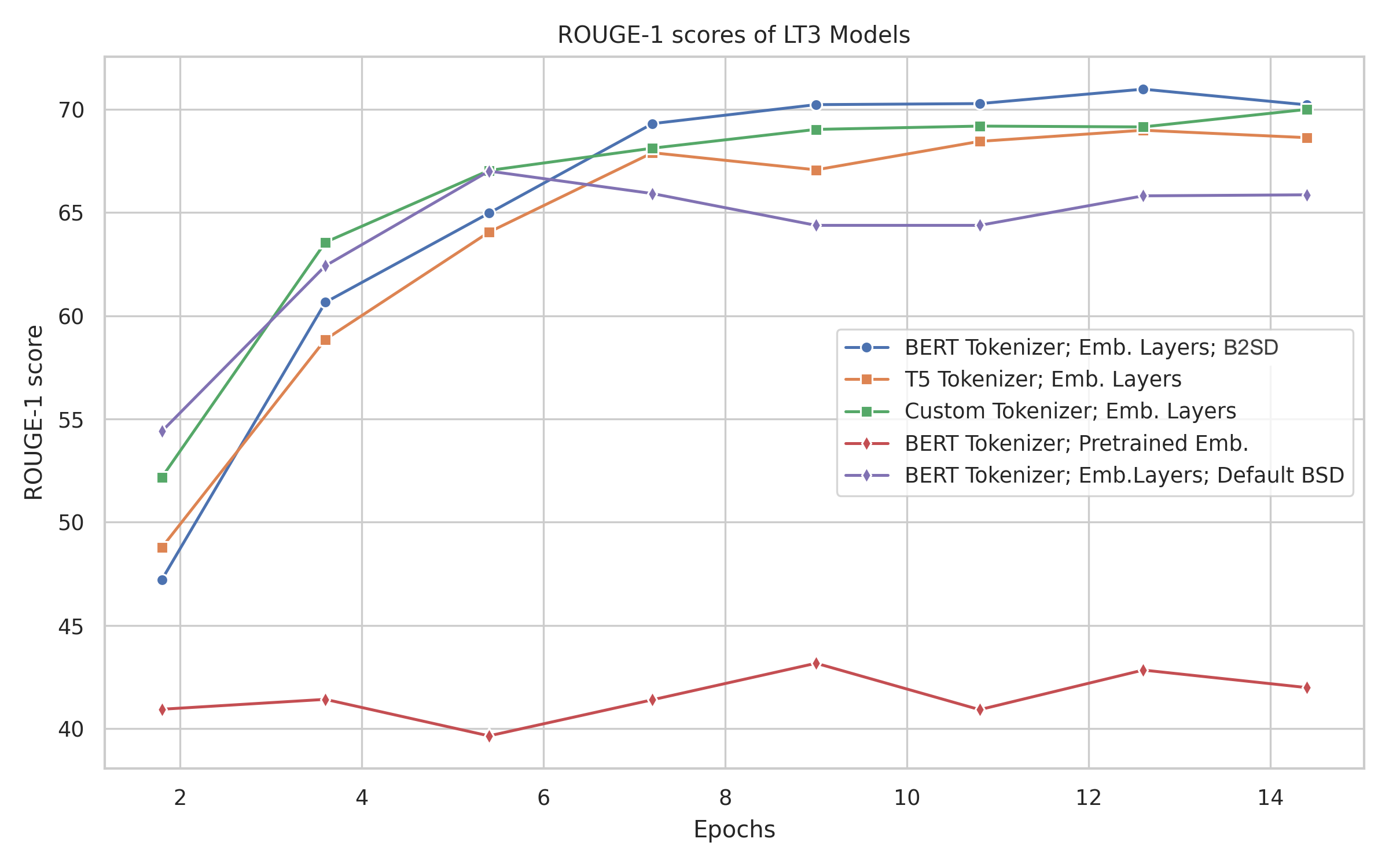}
  \end{subfigure}
  \begin{subfigure}{0.49\textwidth}
    \includegraphics[width=\linewidth]{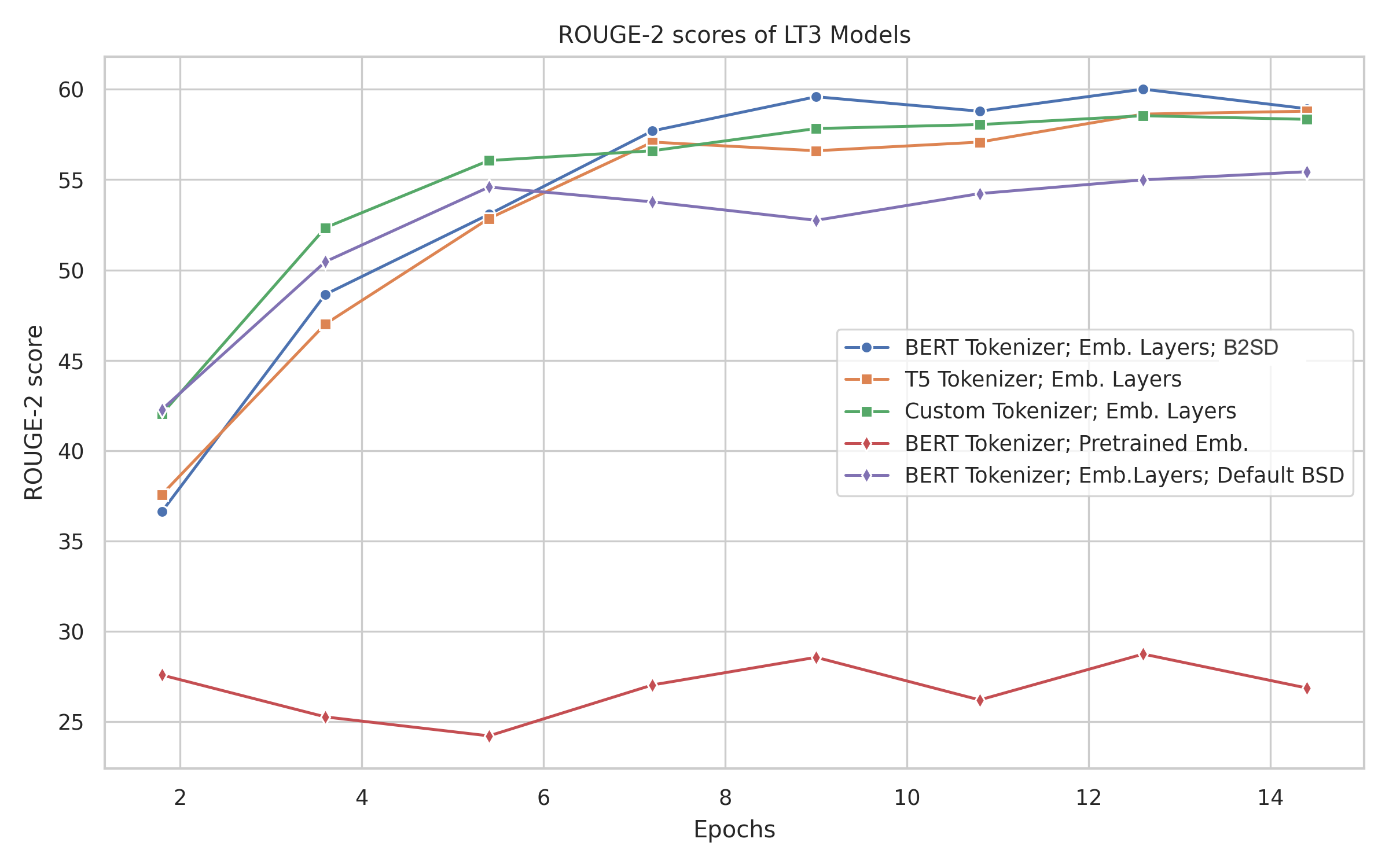}
  \end{subfigure}
  \hfill
  \begin{subfigure}{0.49\textwidth}
    \includegraphics[width=\linewidth]{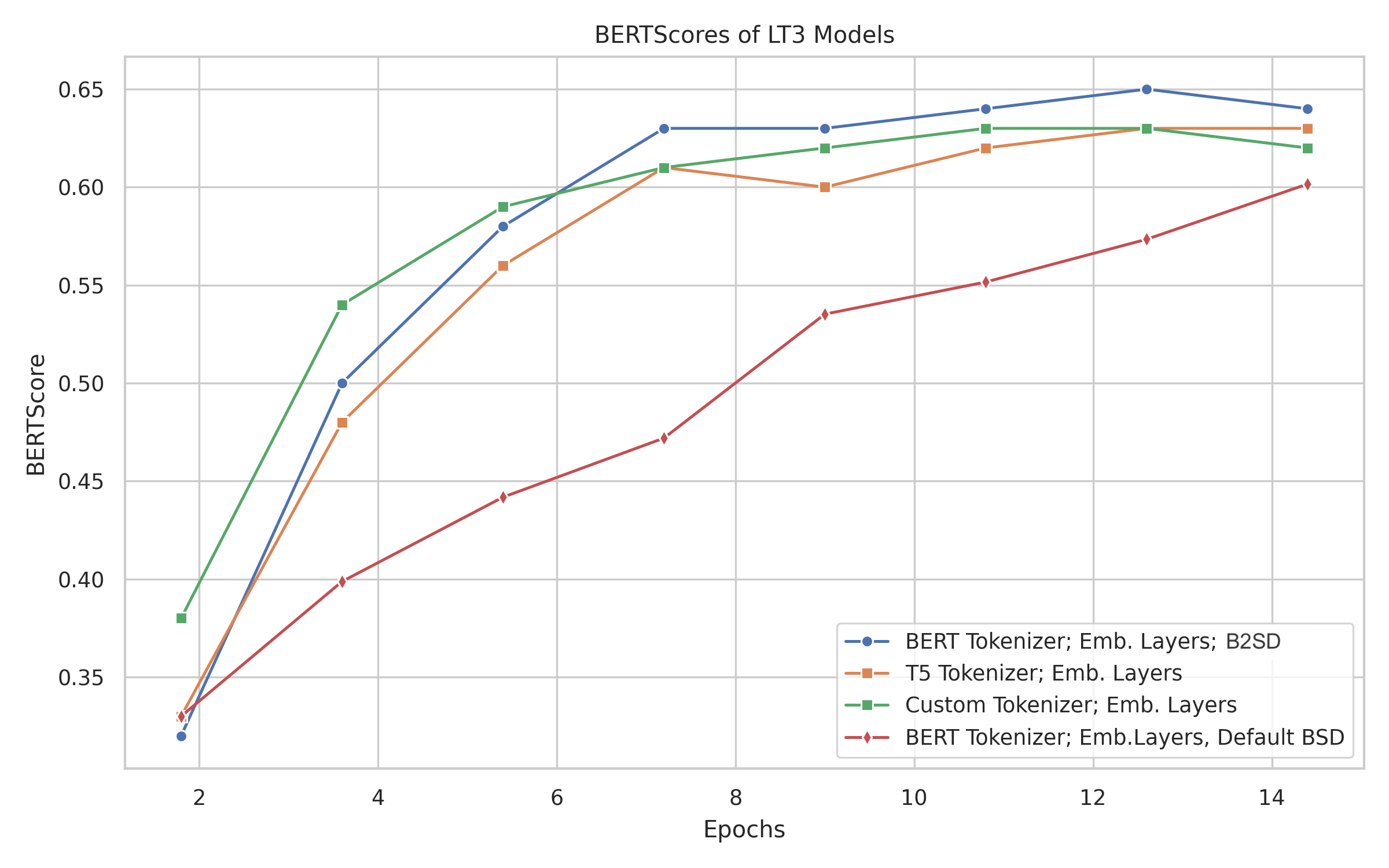}
  \end{subfigure}
  \caption{Quantitative Evaluation Scores of LT3 Models on the Testing Set}
  \label{fig:bsd_compare}
\end{figure}

\section{Comparisons on Beam Search Decoding Algorithms}
\label{appendix:beamsearchcomparisons}
To quantify the difference in execution time between the original BSD algorithm and the proposed backtracking variant, we ran the following experiment on a TPU v2.

Initially, the validation set is 304 samples divided into 157 unique labels, with a median of 36 samples per label. This experiment used LT3 to generate four synthetic datasets from the validation set by increasing its size by 2, 5, 7, and 10. The increase in size is proportional to the number of samples per unique label. Hence, the same number of unique labels remains while the number of samples increases. For instance, if the first label has three samples, it will be increased to 6 in the first synthetic dataset, 15 in the second, etc. Thus, we force the beam search tree to expand in size for each label to quantify its impact on the execution time.

For each synthetic dataset, we use five different versions of the LT3 model from different checkpoints of its training. This is done to simulate the execution time of the algorithm on models of varying efficiency and certainty.

\begin{figure}[]
  \centering
  \begin{minipage}[b]{0.85\textwidth}
    \includegraphics[width=\textwidth]{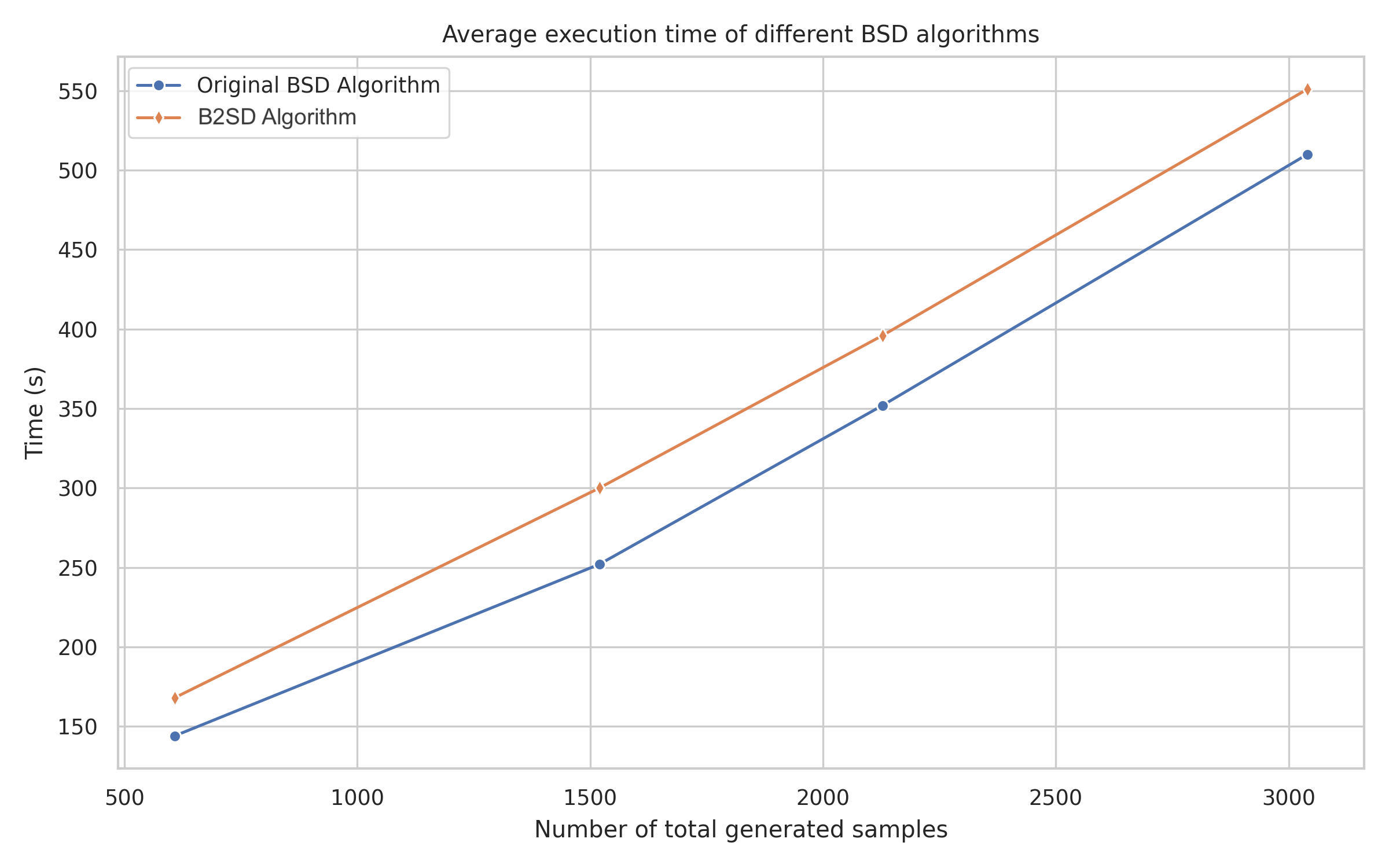}
    \caption{Average Execution Time of Original BSD and B2SD Algorithms}
    \label{fig:bsd_exec_times}
  \end{minipage}
\end{figure}

In practice, we observe a rather linear increase in complexity when using both algorithms, reducing the huge trade-off in their theoretical complexities. LT3 deals with a limited number of samples per generation, and the generated sequences are relatively short. On the other hand, most of the advantages of the backtracking algorithm are preserved.

It is important to note that, whereas B2SD uses a heuristic function based on the joint probability of a sequence, this algorithm will perform the best on well-trained models with certainty in their token selection, meaning high distinction between sequence probabilities. This ensures that the algorithm goes straight at generating the most promising sequences. However, on ineffective or untrained models, it may perform slowly as it might consider many dead-end sequences where probabilities are close to each other due to uncertainty in token generation.

\section{On the Evaluation Settings}
\label{appendix:eval_settings}

\begin{figure}[h]
  \centering
  \begin{minipage}[b]{1\textwidth}
    \includegraphics[width=\textwidth]{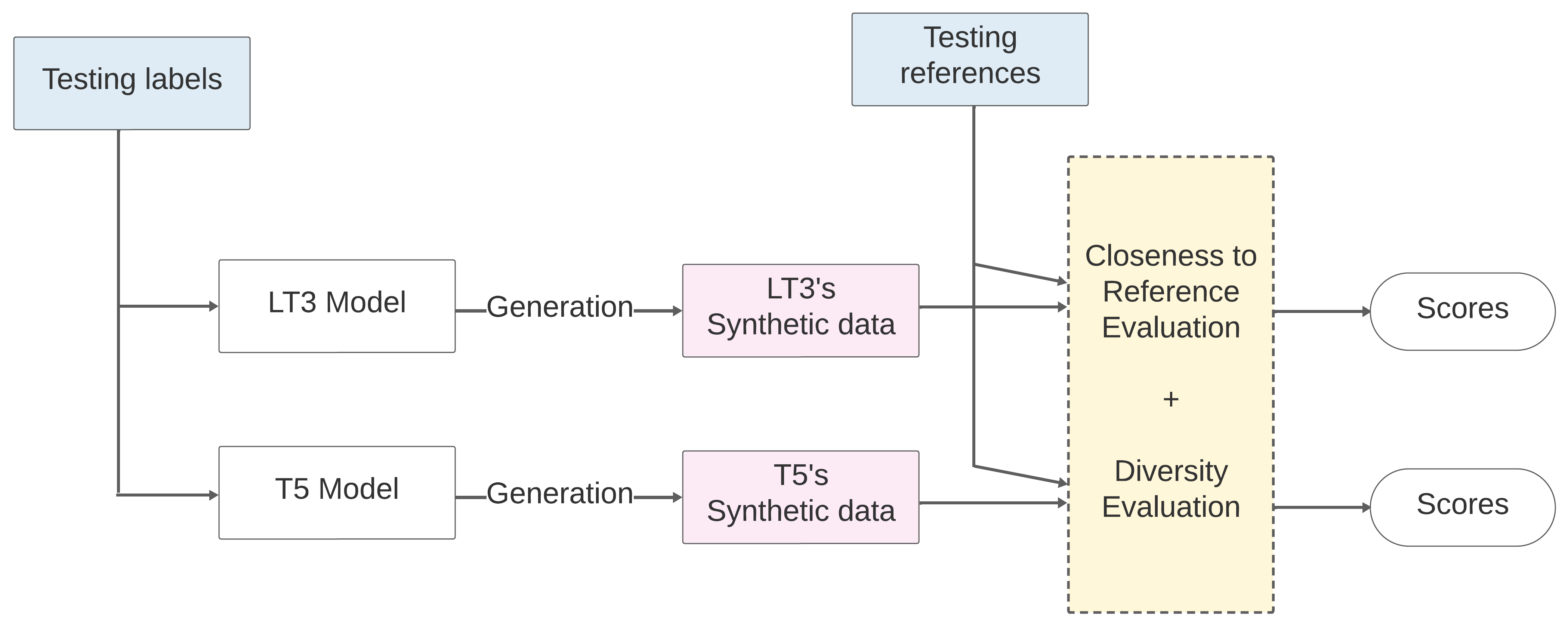}
    \caption{Lexical Evaluation Pipeline}
    \label{fig:quantitative_eval_pipeline}
  \end{minipage}
\end{figure}

To provide a lexical evaluation of the generated data, we aim to assess the performance of LT3 compared to T5-small, T5-base, and T5-large at generating synthetic prescriptions from unseen data (Figure \ref{fig:quantitative_eval_pipeline}). 
To process the comparison, we use the labels from the testing set to generate synthetic data, creating a five times larger dataset than the original testing dataset. For instance, ten prescriptions will be generated if a particular label appears twice in the testing data, ten prescriptions will be generated. We conduct two types of evaluations:

\begin{itemize}
    \item \textbf{Quantitative Evaluations} to assess the quality of the generated prescriptions by comparing LT3 and T5 against reference prescriptions.
    \item \textbf{Lexical Diversity Evaluation} to measure the diversity of the generated prescriptions from LT3 compared to T5.
\end{itemize}

The overall framework of this experimental design for lexical evaluation is displayed in Figure \ref{fig:quantitative_eval_pipeline}.
This experiment aims to show that \textbf{(1)} LT3 can generate lexically diverse prescriptions, as well as \textbf{(2)} significantly larger volume of data compared to the available real data. \textbf{(3)}, despite generating a larger dataset, we intend to confirm that the quality of LT3's generated prescriptions remains high in terms of quantitative scores against references. \textbf{(4)} Most importantly, we try to assess LT3's overall abilities at generating prescriptions from unseen data.

\section{Model hyperparameters}

\begin{tabular}{ |p{3cm}||p{3cm}|p{3cm}|p{3cm}|  }
     \hline
     Parameters& LT3 & T5 Base & T5 Large \\
     \Xhline{2\arrayrulewidth}
     $d_{model}$     & 515   & 768   & 1024    \\
     $d_{ff}$        & 2038  & 3072  & 4096    \\
     $d_{kv}$        & 64    & 64    & 64    \\
     Dropout     & 0.2   & 0.1   & 0.1    \\
     Heads       & 5     & 12    & 16    \\
     Layers      & 2     & 12    & 24    \\
     \hline
     Learning rate & 0.0004 & 0.001  & 0.001 \\
     Weight decay  & 0.02   & 0.02   & 0.02  \\
     \hline
     Epochs        & 10     & 10     & 10    \\
     Batch size    & 53     & 10     & 10    \\
     FP16          &        & False  & False \\
     Optimizer     & AdamW  & AdamW  & AdamW \\
     \hline
     Params (x$10^6$)  & \textbf{57}     & \textbf{220}     & 770    \\
     \hline
\end{tabular}

\vspace{.5cm}
where:
\begin{itemize}
    \item $d_{model}$ represents the dimension of the model's hidden states or embeddings;
    \item $d_{ff}$ represents the dimension of the feed-forward network within the Transformer's self-attention layers;
    \item $d_{kv}$ represents the dimension of the query, key, and value vectors used in the attention computation.
\end{itemize}

\end{appendices}

\begin{comment}

References follow the acknowledgements. Use unnumbered first-level headings for
the references. Any choice of citation style is acceptable as long as you are
consistent. It is permissible to reduce the font size to \verb+small+ (9 points)
when listing the references.
Note that the Reference section does not count towards the page limit.
\medskip

{
\small

[1] Alexander, J.A.\ \& Mozer, M.C.\ (1995) Template-based algorithms for
Connectionist rule extraction. In G.\ Tesauro, D.S.\ Touretzky and T.K.\ Leen
(eds.), {\it Advances in Neural Information Processing Systems 7},
pp.\ 609--616. Cambridge, MA: MIT Press.

[2] Bower, J.M.\ \& Beeman, D.\ (1995) {\it The Book of GENESIS: Exploring
  Realistic Neural Models with the GEneral NEural SImulation System.}  New York:
TELOS/Springer--Verlag.

[3] Hasselmo, M.E., Schnell, E.\ \& Barkai, E.\ (1995) Dynamics of learning and
recall at excitatory recurrent synapses and cholinergic modulation in rat
hippocampal region CA3. {\it Journal of Neuroscience} {\bf 15}(7):5249-5262.
}
\end{comment}

%%%%%%%%%%%%%%%%%%%%%%%%%%%%%%%%%%%%%%%%%%%%%%%%%%%%%%%%%%%%

\end{document}